\documentclass[9pt,a4paper,twoside]{rho-class/rho}
\usepackage[english]{babel}
\usepackage{cleveref}
\usepackage{amsmath}
\usepackage{tikz, pgfplots}
\usetikzlibrary{positioning}
\usetikzlibrary{calc}
\usetikzlibrary{shapes}
\usetikzlibrary{patterns, patterns.meta}

\setbool{rho-abstract}{true} %
\setbool{corres-info}{true} %

\usepackage{graphicx}
\usepackage{array}

\usepackage{listings}
\usepackage{xcolor}
\usepackage{caption}

\usepackage{algorithm}
\usepackage[noend]{algpseudocode}

\title{Benchmarking symbolic regression constant optimization schemes}

\author[1,$\dagger$]{L.G.A dos Reis}
\author[2]{V.L.P.S. Caminha}
\author[3,$\dagger$]{T.J.P.Penna}

\affil[1]{ICEx UFF}
\affil[2]{ICEx UFF}
\affil[3]{ICEx UFF}
\affil[$\dagger$]{These authors contributed equally to this work}

\dates{This manuscript was compile on December 2, 2024}

\leadauthor{L.G.A dos Reis}
\footinfo{Creative Commons CC BY 4.0}
\institution{ICEx UFF}
\theday{December 2, 2024} %

\corres{L.G.A dos Reis}
\email{lreis.luiz@gmail.com}
\license{Rho LaTeX Class \ccLogo\ This document is licensed under Creative Commons CC BY 4.0.}

\begin{abstract}
   Symbolic regression is a machine learning technique, and it has seen many advancements in recent years, especially in genetic programming approaches (GPSR). Furthermore, it has been known for many years that constant optimization of parameters, during the evolutionary search, greatly increases GPSR performance However, different authors approach such tasks differently and no consensus exists regarding which methods perform best. 
   
   In this work, we evaluate eight different parameter optimization methods, applied during evolutionary search, over ten known benchmark problems, in two different scenarios. We also propose using an under-explored metric called Tree Edit Distance (TED), aiming to identify symbolic accuracy. In conjunction with classical error measures, we develop a combined analysis of model performance in symbolic regression. We then show that different constant optimization methods perform better in certain scenarios and that there is no overall best choice for every problem. Finally, we discuss how common metric decisions may be biased and appear to generate better models in comparison.
\end{abstract}

\keywords{Symbolic Regression, Machine Learning, Tree Edit Distance, Constant Optimization, Benchmark}

\begin{document}
	
    \maketitle
    \thispagestyle{firststyle}

\section{Introduction}

Applications of artificial intelligence in many different areas have been standard practice in recent years. Symbolic regression (SR) is one of those rising techniques, with applications that range from materials science \cite{wang2019symbolic}, the energy sector \cite{ANDELIC2024108213}, to psychology \cite{miyazaki2023application} and especially scientific discovery \cite{makke2024interpretable}, in areas such as physics \cite{udrescu2020ai}. SR is a machine learning technique that finds a symbolic expression that best fits a dataset. In the opposite direction from traditional AI methods, symbolic regression represents a white-box approach that searches for a mathematical function that describes a given scenario.

Since its origins, many approaches have been proposed for symbolic regression implementations and improvements. There are many new techniques proposed to model SR, such as those structured around Bayesian statistics \cite{jin2019bayesian}, random search \cite{mcconaghy2011ffx}, or more recently deep neural networks and large language models \cite{landajuela2022unified, d2022deep, bertschinger2023metric}. Nevertheless, the area is still dominated by genetic programming implementations \cite{cranmer2023interpretable, udrescu2020ai, kommenda2013nonlinear, schmidt2010age, koza1994genetic}, called GPSR, since they are more robust, produce better results, and have a way vaster literature than any other alternative. 

Said that, a well-known fact about GPSR models is the impact of constant optimization during the evolutionary process. Many studies have shown a great increase in performance once a technique such as \textit{simulated annealing} \cite{esparcia1997learning}, \textit{particle swarm optimization} \cite{korns2013baseline}, \textit{non-linear least squares} \cite{kommenda2013nonlinear, kommenda2020parameter, topchy2001faster} or \textit{differential evolution} \cite{mukherjee2012differential} has been applied. Other popular models, such as \textit{PySR} \cite{cranmer2023interpretable}, utilize different alternatives, such as optimization by \textit{BFGS} or \textit{Nelder-Mead}. However, even though this information is common sense among symbolic regression researchers, there are no studies that identify which is the best parameter optimization method to use in a model, or the impact different alternatives have on convergence. A discussion about distinct optimization approaches after a solution has been found was developed in \cite{de2015evaluating}, but no study has demonstrated yet the different impact they cause as part of the GP method.

Considering the aforementioned, this paper performs a comprehensive benchmark study on the impact modern constant optimization techniques have on the SR performance, once they are implemented as part of the evolutionary search. We also propose the use of Tree Edit Distance as a metric to understand SR performance, to not only quantify how well a solution fits the data but also how symbolically accurate an expression is. To achieve this, we share a preprocessing algorithm that simplifies expressions to a common denominator, for accurate comparison. 

This study is structured as follows. In \cref{sec:problem_definition} we describe the problem of symbolic regression and constant optimization. The methods studied, the metrics utilized, and the preprocessing technique developed are described in the methodology \cref{sec:methodology}. Results are presented and discussed in \cref{sec:results_and_discussion} and our conclusions are exposed in \cref{sec:conclusions}.

\section{Motivation}
\label{sec:problem_definition}

Even though many different approaches to symbolic regression have been proposed currently, since the start \cite{koza1994genetic} of the area, SR algorithms were constructed upon genetic programming techniques (GPSR). These still represent a great proportion of state-of-the-art models \cite{cranmer2023interpretable,udrescu2020ai,virgolin2021improving,kommenda2020parameter}. In this, mathematical expressions are represented in a computer, usually as expression trees, and are evolved through a process of mutation, crossover, and selection, to better approach the input data.

A well-established concept in GPSR is that of constant optimization. Many studies \cite{burlacu2013evolutionary, kommenda2013nonlinear, de2015evaluating, korns2011abstract} have already shown that by optimizing the parameters of the discovered expression in the evolutionary process, the model performance increases greatly. Not only concerning the minimization of numerical metrics but also by finding the correct mathematical expression that describes the data. For that reason, most methods of GPSR utilize some algorithm to optimize parameters in their model\cite{cranmer2023interpretable, udrescu2020ai, udrescu2020ai2.0, schmidt2010age, virgolin2021improving, kommenda2020parameter}.

\begin{figure}[h!]
	\centering
	\includegraphics[width=\linewidth]{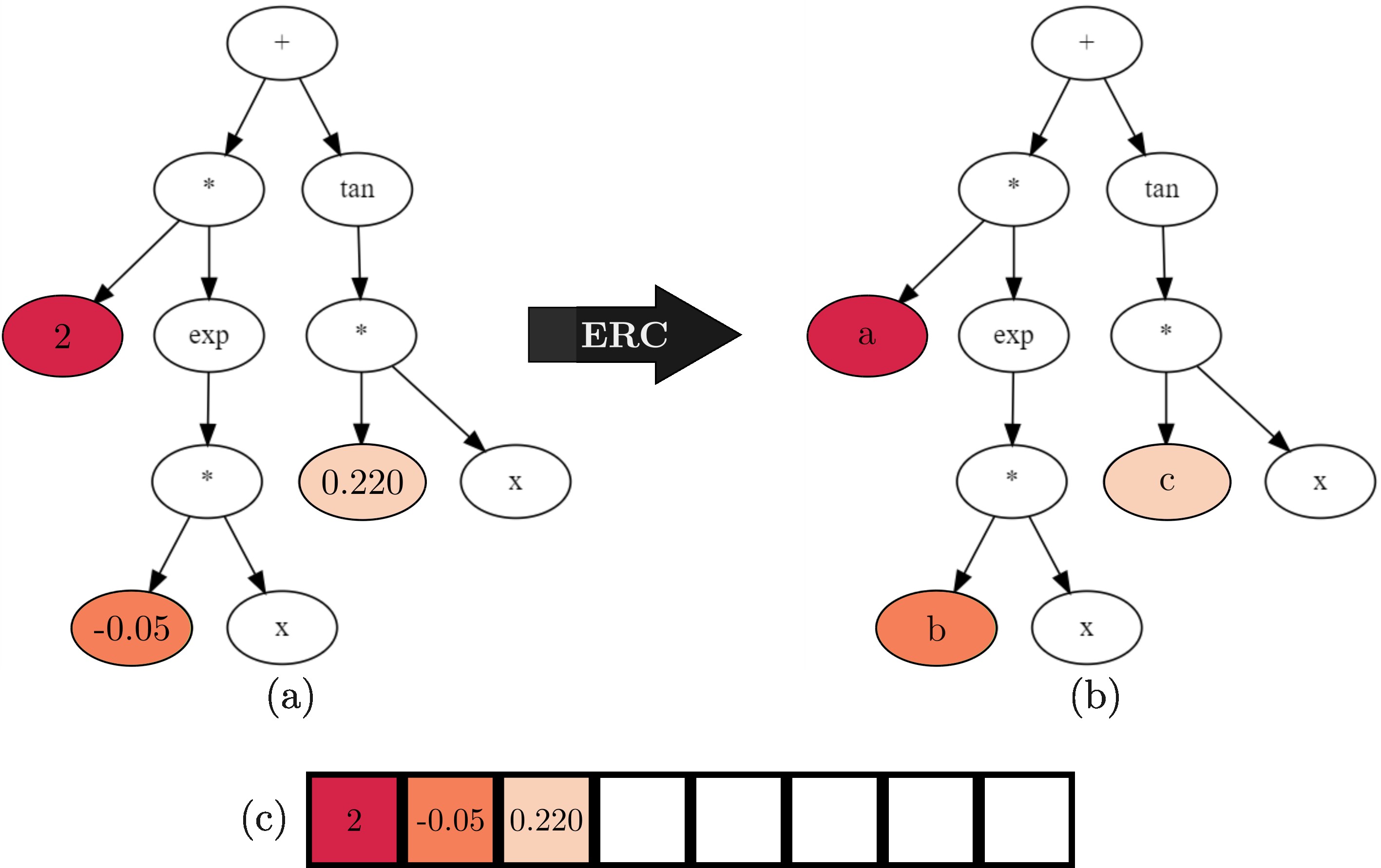}
	
	\caption{Constants (terminal nodes) in (a) expression tree are replaced by (b) ephemeral random constants. (c) is a vector that stores the constants somewhere else for easy access and future manipulation.}
	\label{fig:ERC_transform}
\end{figure}

The process of constant optimization in SR usually takes a common form. The constants of candidate mathematical expression are represented as abstract symbols, that are further set to constant at evaluation time. This general idea presents itself in many different applications, such as Abstract Expression Grammar (AEG) \cite{korns2011abstract, korns2013baseline}, or Ephemeral Random Constants (ERC) \cite{virgolin2021improving}. In tree representation, the terminal nodes (parameters) are replaced with ephemeral constants and a vector stores the current value of each parameter (\cref{fig:ERC_transform}). Through the evolutionary process, the expression changes, the constant optimization method is re-applied, and new constants are generated, which in turn updates the vector of parameters. Once tree evaluation is necessary, it simply requires substituting the stored values back in the ephemeral constants. 

Nevertheless, even though most GPSR methods utilize some algorithm for constant optimization in ERC parameters (\cref{fig:ERC_transform}), there are no studies so far that investigate which methods perform better in this context. Specifically, during the evolutionary process, and in what kind of applications these algorithms excel. To answer this question, we will utilize a straightforward GPSR model, developed by the author and based upon \cite{korns2013baseline}. This was chosen instead of a consolidated state-of-the-art algorithm because these models are usually highly complex and perform many different operations and simplifications, that could interfere in unknown ways with the benchmark. So a simple model, that is composed only of the bare bones of a generic programming algorithm, should highlight the strong suits of each constant optimization method. At the same time, the analysis is general enough to apply to any of the GPSR models currently in the market, since all of them are based on the same core.

\section{Methodology}
\label{sec:methodology}
Following we shortly describe the unconstrained constant optimization algorithms evaluated in the benchmark. Each is applied to the ERC, representing the equation, as shown in \cref{fig:ERC_transform}, and an updated constant vector is generated. Additionally, \textit{BFGS}, \textit{CG}, \textit{LS}, and \textit{Nelder-Mead} require an initial guess, for these methods, we tested two different situations. Firstly, we insert back the current constants vector as an initial guess, this case is stated in the tests simply as the optimizations name. Secondly, we generate a random vector sampled from a normal distribution as an initial guess. Optimizations treated this way are followed by \textit{Random}, after their names. This was done to understand the impact an initial guess would have on the capability of such methods to explore the search space. 

\begin{enumerate}
	\item \textbf{\textit{NoOpt}}
	
	No Optimization (NoOpt) represents the dummy method, where no constant optimization algorithm is applied. It serves as a baseline and comparison to understand how effective constant optimizations are.\\
	
	\item \textbf{\textit{Broyden-Fletcher-Goldfarb-Shanno - BFGS}}
	
	Quasi-Newton methods are a class of optimization algorithms that use updating formulas to approximate the hessian of a differentiable function, and from that solve unconstrained optimization problems. The hessian matrix is utilized to guide the search for a minimum. BFGS has been proved to be the most effective of those methods, and it utilizes gradient evaluations to update the hessian approximation, without explicitly calculating second derivatives. Applications range from engineering and physics to machine learning. \cite{dai2013perfect}\\
	
	\item \textbf{\textit{Conjugate Gradient Descent - CG}}
	
	Gradient Descent is one of the most well-known optimization methods for minimizing unconstrained nonlinear functions and solving systems of linear equations. Multiple versions inspired by the original algorithm were developed, but conjugate gradient descent has shown faster convergence than other alternatives. It works by moving in the direction generated by combining the steepest descent and the conjugate direction. Recently, it has been widely used in deep learning algorithms, particularly in the context of high-dimensional data. \cite{nazareth2009conjugate}\\
	
	\item \textbf{\textit{Levenbergâ€“Marquardt Least Squares - LS}}
	
	Levenberg-Marquardt is a common optimization approach to solve the least squares minimization problem. It combines the Gauss-Newton method and the already discussed gradient descent. The LM algorithm adaptively changes between them by considering a damping coefficient $\lambda$. Such coefficient is initialized to be large, which results in gradient descent steps in the downhill direction. Once the solution improves, $\lambda$ is reduced, and the Gauss-Newton method enters into play, to effectively reach the minima. Levenberg-Marquardt is commonly applied in curve fitting for functions with many parameters, but it also sees use in machine learning and data fitting. \cite{gavin2019levenberg}\\
	
	\item \textbf{\textit{Particle Swarm Optimization - PSO}}
	
	Particle Swarm Optimization is a method inspired by the flocking of birds. Potential solutions are modeled as particles that move through the search space. Each coordinate of such particles in this space represents a parameter to be optimized. Positions evolve by calculating a velocity vector that depends on three pieces of information: the best position each has encountered so far, the best position found by any particle in the system, and an inertia vector. PSO is particularly useful for high-dimensional, nonlinear problems and is suited for applications where gradients cannot be calculated. \cite{kennedy2006swarm}\\
	
	\item \textbf{\textit{Nelder-Mead}}
	
	Nelder-Mead is an optimization algorithm that works by adapting a geometric shape called simplex, which consists of $n+1$ vertices in a $n$-dimensional space. It is independent of derivative calculations, where the simplex is modified by processes of reflection, expansion, contraction, and shrinkage. At each iteration, the model explores the search space by function evaluation on each vertex. It is widely used in real-world situations for optimizing functions where the gradient is unknown. The method also works well for smooth, unimodal problems but its efficiency may suffer in high-dimensional landscapes. \cite{gao2012implementing}\\

	\item \textbf{\textit{Differential Evolution}}
	
	Differential evolution (DE) is a simple population-based, stochastic, evolutionary algorithm. In some comparisons, DE has presented greater efficiency than many stochastic methods, such as simulated annealing, evolutionary programming, and particle swarm optimization. It evolves a population of candidate solutions by processes of mutation, crossover, and selection, similar to GPSR. Differential evolution is simple yet effective in exploring large, multi-modal search spaces. For that reason, it is sometimes applied in global optimization tasks for which traditional methods struggle.\cite{qiang2014unified}\\
	
	\item \textbf{\textit{Generalized Dual Annealing}}
	
	Generalized Dual Annealing (GSA), also called dual annealing by the \textit{scipy} implementation (and the name we use), is an optimization algorithm inspired by the annealing process in metallurgy. Classical simulated annealing (CSA) creates a modified version of the current solution every iteration and the probability of acceptance for this new solution is a function of "temperature". This parameter starts very high, to explore more space, and then cools off. CSA was then combined with Fast Simulated Annealing to create GSA. Dual annealing is adapted to complex, nonlinear, and multi-modal problems, it is widely used in combinatorial optimization and engineering. \cite{xiang2013generalized}
	
\end{enumerate}

\subsection{Test Problems and Search Space}

To study the behavior of multiple optimization methods, a comprehensive set of test problems was selected accordingly to \cref{tab:problems}. Only univariate functions were chosen to better understand the effects of the optimization methods in similar problems - the difficulty comes from the number of constants and complex relationships, not extra dimensions. 

\bgroup
\newcolumntype{L}{>{\raggedright\arraybackslash}p{1.5cm}}
\newcolumntype{M}{>{\raggedright\arraybackslash}p{1.5cm}}
\begin{table}[h!]
	\centering
	\begin{tabular}{lLllM}
		\toprule
		& Reference Name & Expression & Interval & Specific Set\\
		\midrule
		0 & F1 & $1.57 + 24.3x$ & [-5, 5] & $x^2$\\[2mm]
		1 & F4 & $-2.3 + 0.13\sin(x)$ & [-5, 5] & $\sin(x)$, $cos(x)$\\[5mm]
		2 & F5 & $3 + 2.13\ln(x)$ & [0.1, 10] & $\ln(x)$ \\[2mm]
		3 & F6 & $1.3 + 0.13\sqrt{x}$ & [0.1, 10] & $\sqrt{x}$\\[2mm]
		4 & F7 & $213.81(1-e^{-0.547237x})$ & [0.1, 10] & $e^{x}$, $e^{-x}$\\[2mm]
		5 & F11 & $6.87 + 11\cos(7.23x^3)$ & [-5, 5] & $\cos(x)$, $\sin(x)$, $x^3$\\[5mm]
		6 & Logistic & $10e^{-0.5e^{-0.5 + 2}}$ & [-5, 15] & $e^{x}$, $e^{-x}$\\[2mm]
		7 & Projectile Motion & $6x - 9.8x^2$ & [-5, 5] & $x^2$\\[5mm]
		8 & Damped Pendulum & $e^{-x/10}(3\cos(2x))$ & [-5, 5] & $e^x$, $e^{-x}$, $\cos(x)$, $\sin(x)$\\[10mm]
		9 & Radioactive Decay & $10e^{-0.5x}$ & [-5, 5] & $e^x$, $e^{-x}$ \\[2mm]
		\bottomrule
	\end{tabular}
	\caption{The problems starting with `F' were extracted from the Korns benchmark functions \cite{mcdermott2012genetic, de2015evaluating}. Every problem was evaluated with 1000 points in the respective interval. The last column refers to the set of functions utilized in each problem's specific case.}
	\label{tab:problems}
\end{table}
\egroup

Moreover, symbolic regression problems are highly dependent on the set of functions utilized as a basis for the search space. In most GPSR methods, one may choose a set of functions and mathematical operators, such as $\{\sin, \cos, \sqrt{}, \ln\}$ and $\{+, *\}$, respectively, to utilize. The algorithm will then sample from these sets the terms that will compose any expression. In other words, this set is composed of the basis functions for the function space that represents the search of the SR model. Any expression that the model can generate is a linear combination of the set of basis functions. Therefore, by increasing the number of elements in this set, the search space increases drastically, which implies a trade-off between solution diversity, i.e., the number of different solutions the model can reach, and the complexity of the regression problem. If the search space is too large, the model may get stuck in local minima, and never find the target solution.

Given what was discussed, we have studied different subsets of the expressions in \cref{tab:problems}, according to the basis functions utilized. The \textit{standard} problems were evaluate with the same set: \{$+$, $-$, $*$, $/$, $x^2$, $x^3$, $\sqrt{}$, $\sin$, $\cos$, $\tan$, $\tanh$, $\text{abs}$, $\ln$, $e^x$, $e^{-x}$\}. Each expression has a corresponding set of specific basis functions shown in table \ref{tab:problems}. Since we are decreasing the complexity of the search task, they can be interpreted as one of the many ways to give information to the model, and thus make the solution easier to find. 

For simplicity, the problems \textit{projectile motion}, \textit{radioactive decay}, \textit{damped pendulum}, and \textit{F11} are not shown in the following analysis. Given that the first two exhibit the same behavior as \textit{F1}, and the other two as \textit{F7}, for every study performed. Complete results can be accessed in the complementary material. 

\subsection{Metrics}

The comparison between optimization methods is calculated through the following metrics.\\

\textbf{Complexity}\;\; Even though many different complexity measures have been proposed for symbolic regression algorithms, we have opted for a simple approach. We calculate it as the trees size, defined as the number of nodes it has. \\

\textbf{Numerical Accuracy}\;\;\; We access the solution's numerical accuracy by two different metrics. First, we consider the mean squared error as a more direct measure of the model's error, defined as
\begin{align}
	\text{MSE} = \frac{1}{n}\sum_{i=0}^n (y_i - \hat{y}_i)^2.
\end{align}
Secondly, we utilize the standard metric for SR benchmarks, the coefficient of determination, defined as
\begin{align}
	R^2 = 1 - \frac{\sum_{i=0}^n (y_i - \hat{y}_i)^2}{\sum_{i=0}^n (y_i - \overline{y}_i)^2}.
\end{align}
Both metrics were chosen to best achieve an equilibrium, since $R^2$ usually favors overfitted models, while $\text{MSE}$ penalizes outliers. \\

\textbf{Symbolic Accuracy}\;\;\; Most current symbolic regression validation is done by one method of numerical accuracy or some variation of a success metric, in which one counts the number of times the solution has reached the exact expected expression. The problem with that analysis is that the first one does not consider how the actual solution looks, which is nonsensical since the purpose of symbolic regression algorithms is to find a mathematical expression. Even though the second analysis considers algebraic representation, it is only a simple and direct measure. No information can be obtained about how close the solution was to the expected expression. 

\begin{figure*}[h!]
	\centering
	\includegraphics[width=0.8\textwidth, trim={ 0 150 0 100 }, clip]{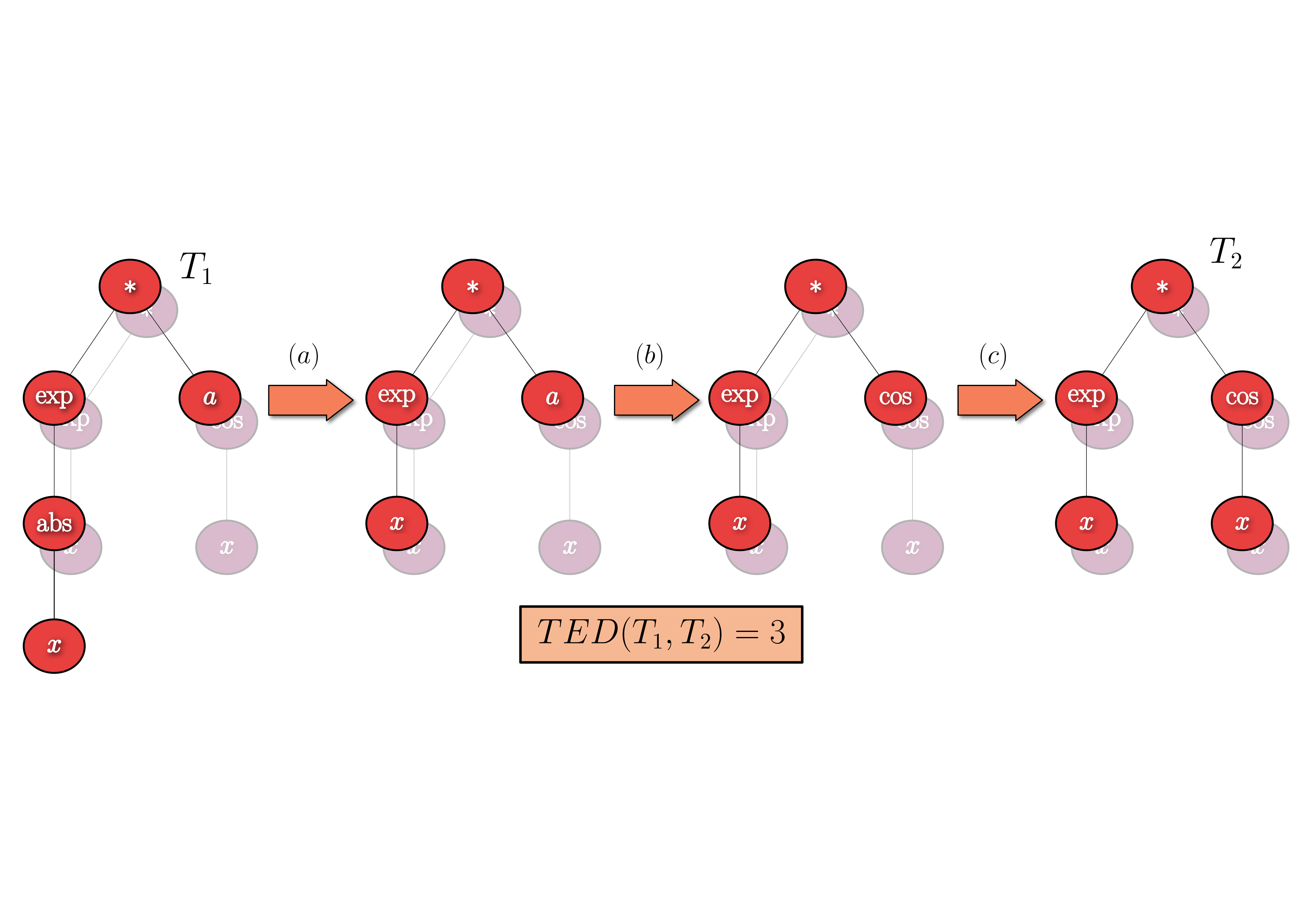}
	\caption{Illustration of the process of TED calculation for an expression tree. The shadow tree represents the expected solution. In the SR context, $T_1$ is the predicted expression, while $T_2$ is the expected solution. (a) The node $\text{abs}$ is removed. (b) $a$ is substituted by $cos$. (c) A variable is added below the $cos$ node. Since three operations were required to transform the $T_1$ into $T_2$, $TED(T_1, T_2) = 3$. }
	\label{fig:TED_visualization}
\end{figure*}

To address these problems we propose the use of Tree Edit Distance ($\text{TED}$), as an equivalent of standard numerical accuracy metric for symbolic closeness. In a recent paper \cite{bertschinger2023metric} has also used $\text{TED}$, alongside other metrics, to study neural symbolic regression. TED is defined as the number of operations of \textit{insertion}, \textit{removal}, and \textit{substitution} necessary to transform a tree into another one. \cref{fig:TED_visualization} shows an example of the process of $\text{TED}$ calculation. \\

\subsection{Preprocessing}

It is common knowledge that for accurately estimating if the model has produced the expected symbolic expression a simplification process must be applied to them. Ordinarily, most authors \cite{bertschinger2023metric, la2021contemporary} use the \textit{simplify} method from Python's \textit{sympy} library. As recognized by some \cite{la2021contemporary}, this is a far too simple process, and many complicated solutions are not correctly evaluated, given that there are many ways to write the same expression.

\begin{figure}[h!]
	\centering
	\includegraphics[width=0.8\linewidth]{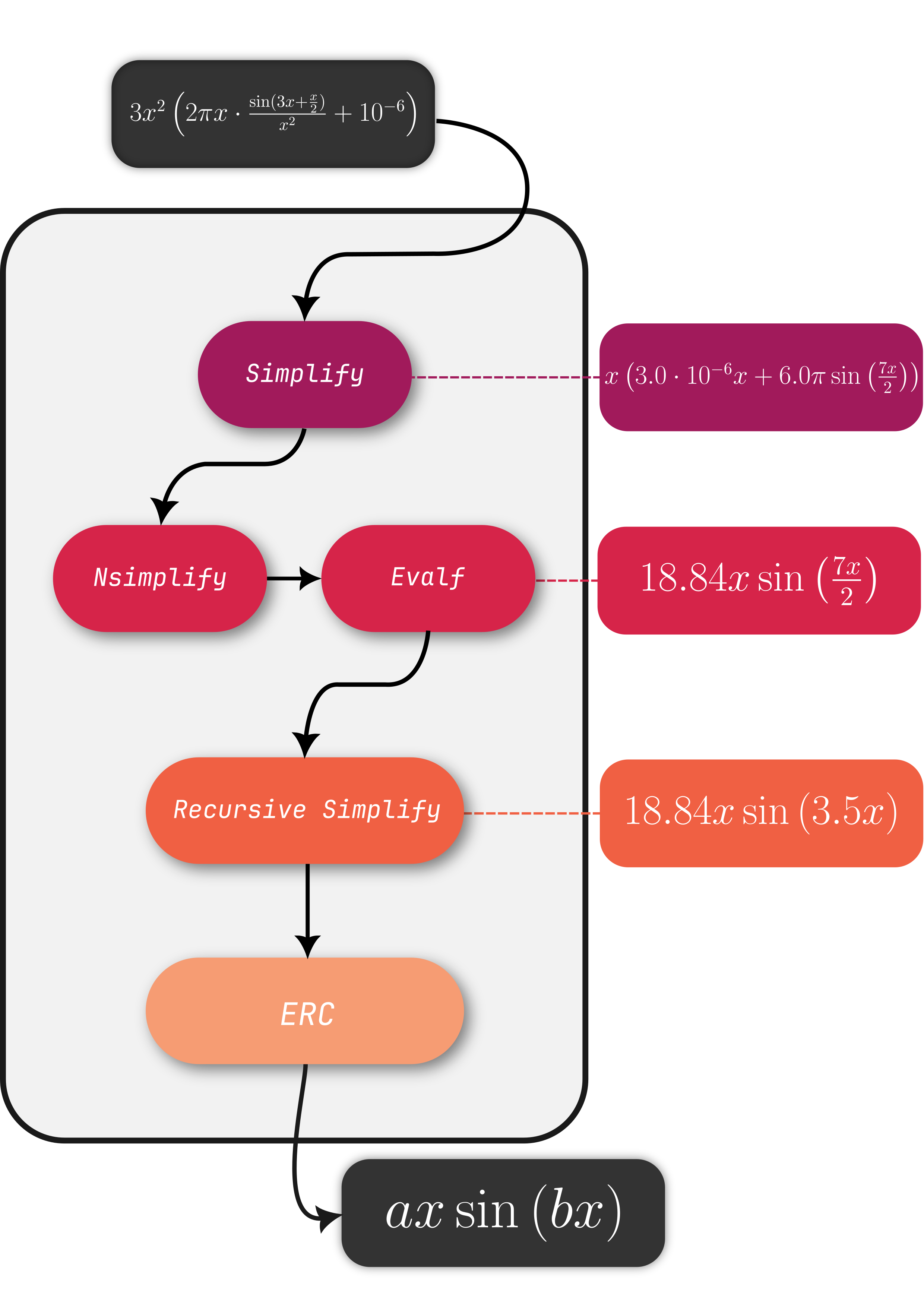}
	\caption{Preprocessing output expression. \textit{Simplify}, \textit{NSimplify} and \textit{Evalf} are methods from Pythons library \textit{Sympy}. \textit{Recursive simplify} is a custom made function showed in code \ref{code:recursive_simplify}. At last, \textit{ERC} represents the conversion of numerical constants into abstract.}
	\label{fig:simplify_config}
\end{figure}

To address the said issue, before calculating $\text{TED}$, every solution underwent a simplification process according to \cref{fig:simplify_config}. Such a procedure is necessary since SR models produce solutions that can be expressed in many different ways. Thus, for an accurate evaluation, all of them must go through a unification process. Also notice that the last step is ERC, given that it is more relevant to recognize the solutions representation instead of how well the parameters are optimized. Once the right expression has been found, any good enough optimization algorithm can work on it and accurately optimize its parameters to fit the data.

The preprocessing starts with a \textit{simplify} to level off major aspects of the expression. Then, \textit{Nsimplify} and \textit{Evalf} are applied in conjunction to remove negligible constants, for which a precision must be defined ($15$ in this study), and simplify fractions and other operations into the minimum number of constants. After that, the custom-made function \textit{Recursive Simplify} \cref{code:recursive_simplify} applies the same simplification procedure in every sub-expression. Lastly, all constants are removed and replaced by abstract representatives by ERC.

	\begin{lstlisting}[language=Python, breaklines=true, numbers=left, tabsize=4, backgroundcolor=\color{black!3}, caption={Python code for recursive simplify function.}, label={code:recursive_simplify}]
def recursive_simplify(expr, precision=15):
	# If the expr has no args, it is an atom (number or symbol)
	if not expr.args:
	return expr.evalf(precision).simplify()
	
	# Recursively process all arguments
	simplified_args = [recursive_simplify(arg, precision) for arg in expr.args]
	
	# Reconstruct expression with simplified arguments
	simplified_expr = expr.func(*simplified_args)
	
	# Remove and Simplify constants
	return simplified_expr.evalf(precision).simplify()
	\end{lstlisting}

\section{Results and discussion}
\label{sec:results_and_discussion}

\begin{figure}[h]
	\centering
	\includegraphics[width=\linewidth]{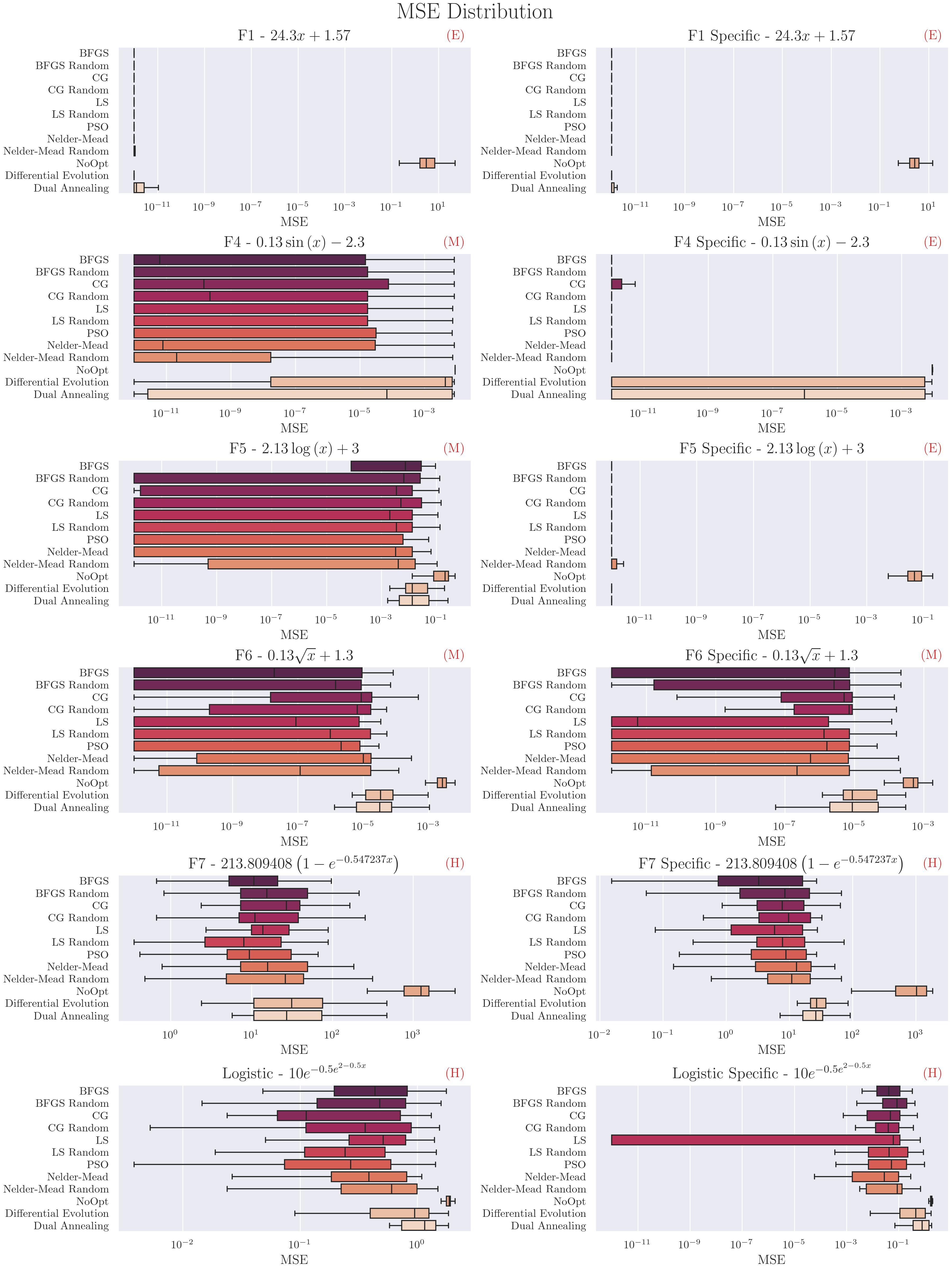}
	\caption{Representative MSE distributions for the problems at hand. Outliers were removed for better visualization. The annotations on the top right corner of each figure are an arbitrary classification used to simplify the analysis, it translates: (E) easy, (M) medium, (H) and hard.}
	\label{fig:MSE_boxplot}
\end{figure}

In \cref{fig:MSE_boxplot} we study the mean squared error distribution for the most representative cases in \cref{tab:problems}. The problems can be classified into three groups, easy, medium, and hard as shown in each figure's top right corner. Such classification helps understand the collective behavior of problems, even though the grouping is arbitrary, and used only to simplify analysis. First, SR without optimization does not converge in any case, as observed for the \textit{NoOpt} control case. It is also noticeable that the Dual Annealing and Differential Evolution methods presented meager results, consistently producing errors significantly greater than every other method. LS has been shown to reach more accurate solutions than other methods, which could imply it is a more robust algorithm to optimize constants precisely.

Secondly, easy cases are independent of the optimization method, and as long as you have one, the model converges with high precision. For medium problems, on the other hand, the result's success is not so easily defined. In cases such as F4, some methods show lower error distributions, but the difference between $10^{-11}$ and $10^{-9}$ is not usually meaningful in most scenarios, and both results could be acceptable. In other problems, the general solution MSE ranges from $10^{-6}$ to $10^{-2}$. Such interval on the error requires specific analysis for the problem at hand, in some scenarios these values could be useless, while in others it could be more than enough. These considerations imply that by looking at a single metric such as numerical error, a straightforward answer regarding the performance of each method is not so easily defined. Now for hard problems, we see a behavior that appears not influenced by constant optimization algorithms, given that no method converged consistently on low error regions. This could imply that the lack of precision is caused by a poor symbolic regression model, which is incapable of finding the right solution, despite the optimization chosen.

\begin{figure}[h!]
	\centering
	\includegraphics[width=\linewidth]{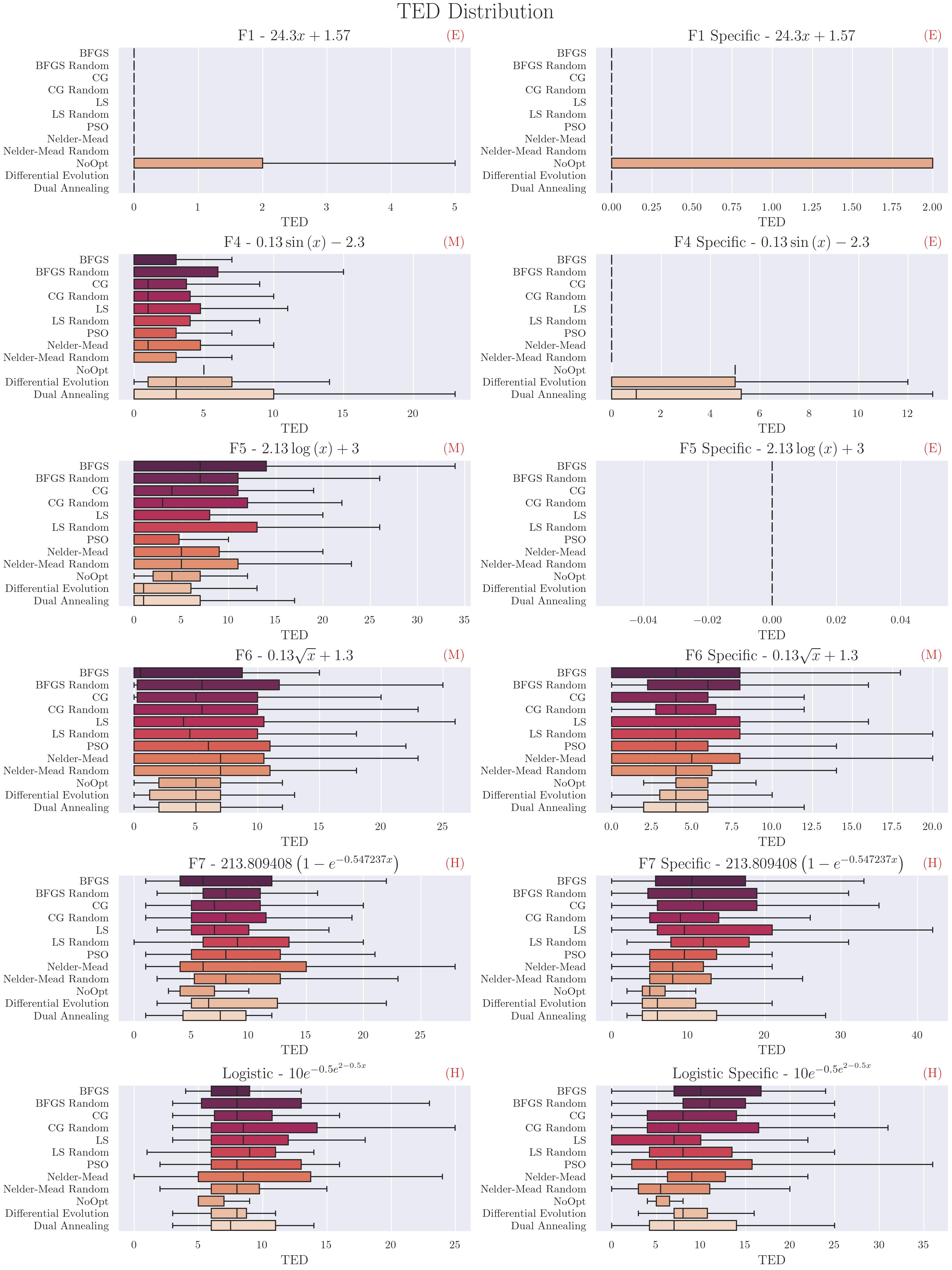}
	\caption{TED distribution for a few representative cases studied. Outliers were removed for better visualization and annotations on the top right corner are an arbitrary classification used to simplify analysis. It reads: (E) easy, (M) medium, and (H) hard.}
	\label{fig:TED_boxplot}
\end{figure}

In \cref{fig:TED_boxplot} we show the TED distribution for the same cases studied. For easy problems, the observed behavior is very similar to \cref{fig:MSE_boxplot}, far too simple problems reach TED of zero, independently of the optimization algorithm (as long as you have one). The only exception is \textit{Dual Annealing} and \textit{Differential Evoltion} for the F4 specific problem, which could indicate that these methods need particular parameter adjustments for each scenario, to perform appropriately. On the other hand, medium cases show a richer behavior, where there are noticeable differences regarding each of the optimization methods. \textit{BFGS}, \textit{LS}, \textit{PSO} and \textit{Nelder-Mead} algorithms show each a problem where they have performed better than average, such as \textit{BFGS} on F6, \textit{LS} and \textit{PSO} on F5, and \textit{BFGS}, \textit{PSO} and 
\textit{Nelder-Mead} on F4. 

Concerning hard problems, two peculiar behaviors emerge. The previously discussed aspect related to the four mentioned algorithms is still present in harder problems, for each scenario one produces better results than the others, even though the variations are smaller. Meanwhile, \textit{DE}, \textit{DA}, and especially \textit{NoOpt} show the oddest behavior. They appear to achieve better TED values than any method, even though they haven't reached significant MSE values, nor presented consistent TED results for easy and medium problems. 

\subsection{Size Correlation}

\begin{figure}[h!]
	\centering
	\includegraphics[width=\linewidth, trim={50 0 50 50}, clip]{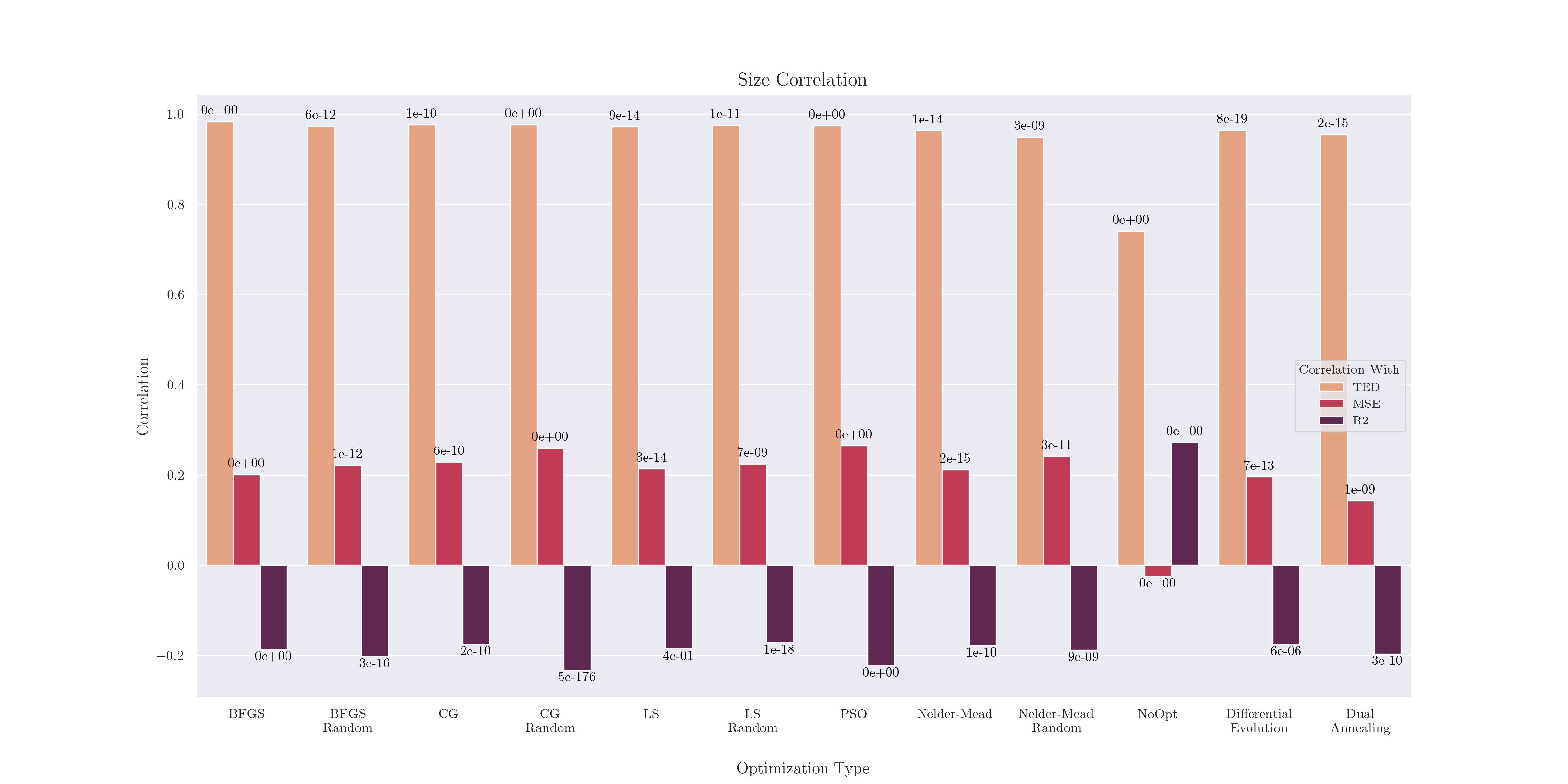}
	\caption{Pearson's correlation between expression size and TED/MSE, for every optimization method discussed. P-values are shown above each bar.}
	\label{fig:size_correlation}
\end{figure}

To better understand the odd $\text{TED}$ behavior, \cref{fig:size_correlation} shows the correlation between TED and MSE with expression size, calculated as the length of the equivalent binary tree. First, we notice that MSE and size are not meaningfully correlated, as one would expect. On the other hand, TED and size have a very high correlation and small p-values for every method, which implies that short expressions produce lower TED values.  

\begin{figure}[h!]
	\centering
	\includegraphics[width=\linewidth]{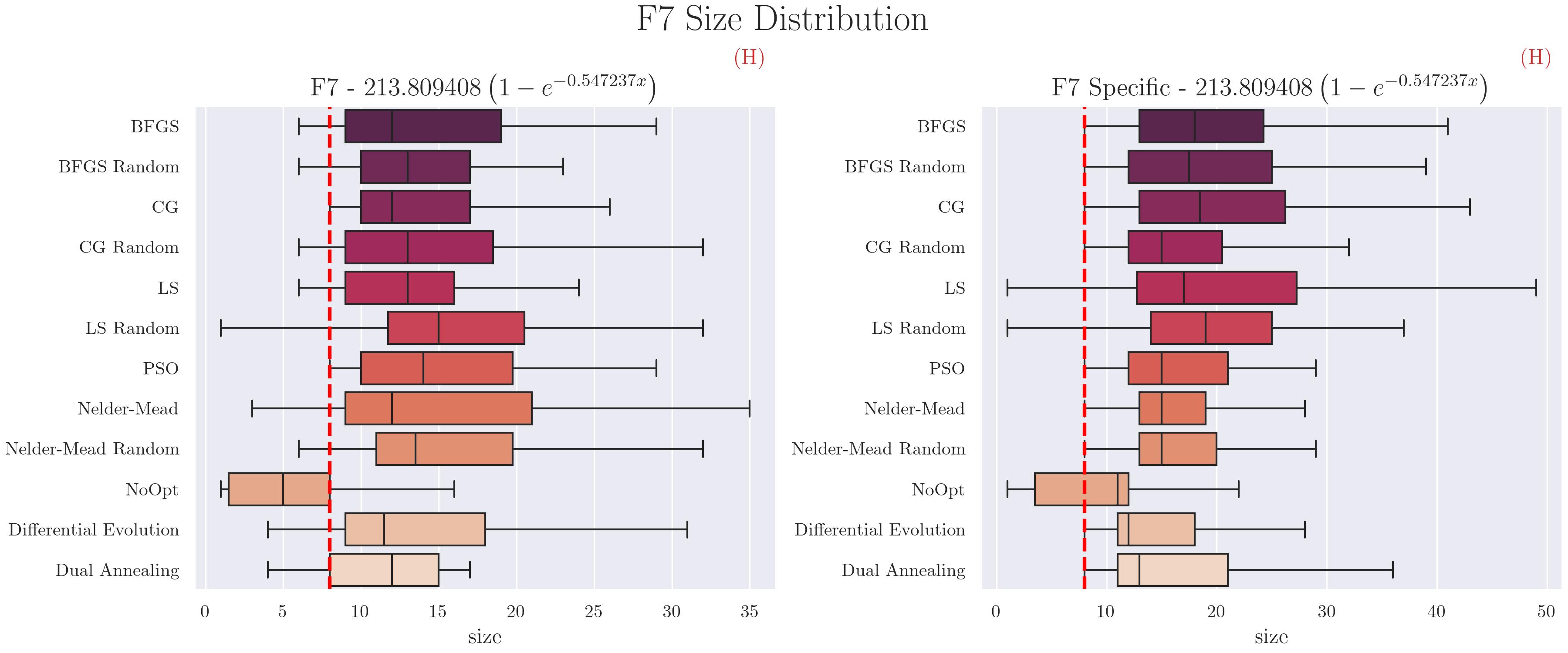}
	\caption{Size distribution for the problem F7 and F7 Specific, the vertical dotted red line represents the expected expression's size. This case serves as an example of common behavior for hard problems.}
	\label{fig:size_distribution}
\end{figure}

Since symbolic regression algorithms try to minimize numerical error, one would expect that if a method isn't capable of converging to the right expression, it would fall into local minima and optimize it the best it can. Such a process would bloat the solution to reduce MSE. Now, if a constant optimization algorithm isn't capable of accurately minimizing the error for new constants, adding more won't increase MSE. Thus the solution stays short, as can be observed in \cref{fig:size_distribution} for the size distribution of \textit{NoOpt}. This combined behavior explains why \textit{NoOpt}, \textit{DA} and \textit{DE} show unexpectedly good results for the TED distribution in figure \cref{fig:TED_boxplot}. Smaller expressions require fewer steps to transform into another since one needs only to create the missing link, not remove unnecessary branches. 

\begin{figure*}[ht]
	\centering
	\includegraphics[width=\linewidth]{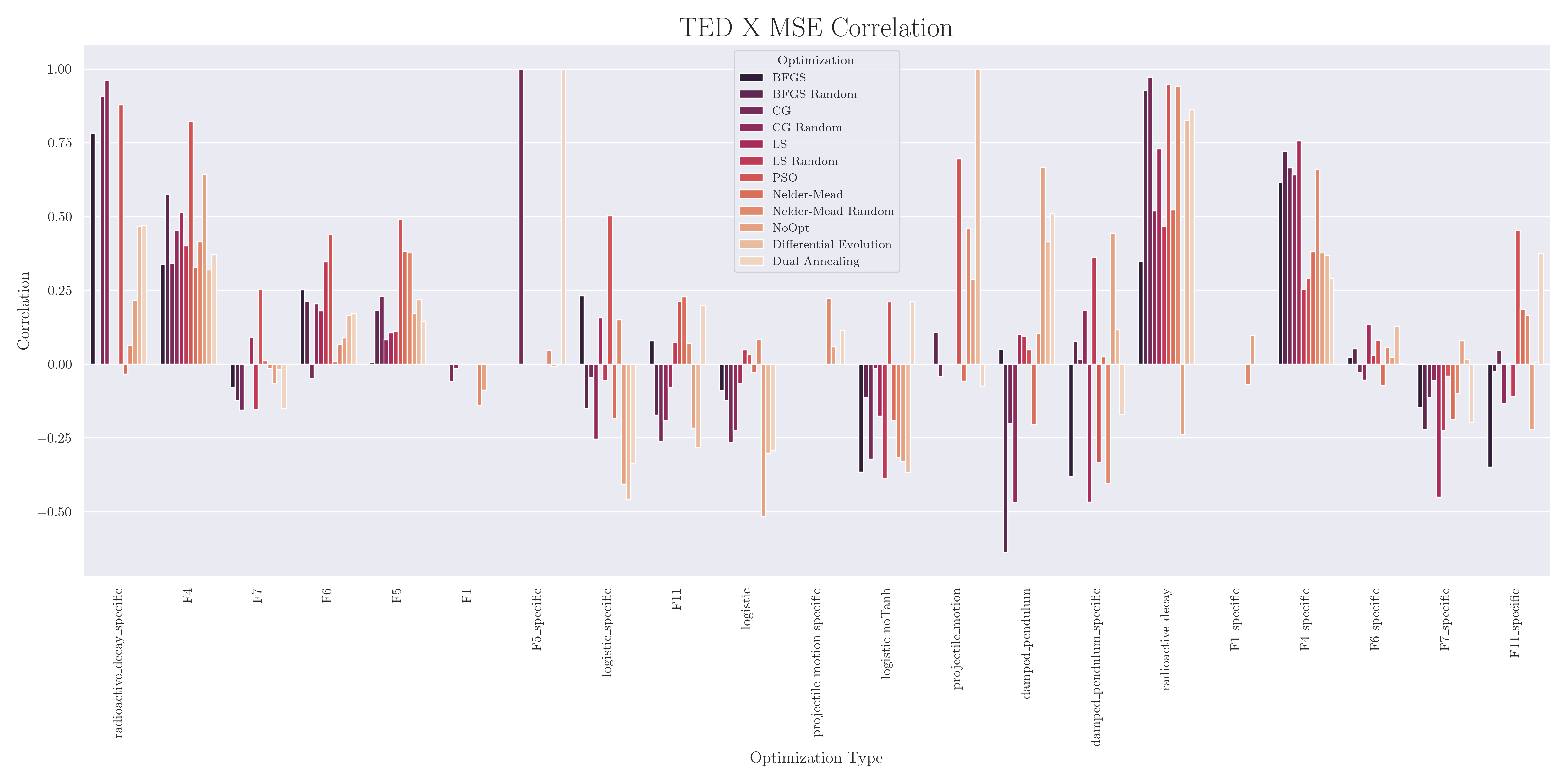}
	\caption{Pearson's correlation between TED and MSE for each optimization algorithm and problem studied. It appears that TED and MSE have no well-defined correlation, and it is optimization and problem-dependent. Both metrics carry unique information.}
	\label{fig:TED_MSE_Correlation}
\end{figure*}

It is important to clarify that we have only considered \textit{NoOpt}, \textit{DA} and \textit{DE} as products of small size distributions for an explanation of abnormal good TED results, because they are the only methods that have presented awful MSE values and greater than average TED for harder scenarios. Following, it is not necessary to develop a new metric that removes the effect of size in TED studies, since this behavior is integral to the problem at hand, given that it will depend on the size of the target expression, which is unknown beforehand.

Furthermore, looking back at \cref{fig:MSE_boxplot} and \cref{fig:TED_boxplot}, we notice that some medium problems reach considerable MSE results, despite showing unsatisfactory TED outcomes. Such cases present very spread-out TED distributions, centered at high values. Therefore, it is logical to assume that the relationship between TED and MSE is not straightforward, and as one may see in \cref{fig:TED_MSE_Correlation}, their correlation is not well defined. 

\subsection{Combined Analysis}

Thus, we conclude that just like one cannot look at MSE alone, TED should not be treated independently, and complementary analysis of size and error should be considered. To address this dependency problem in SR metrics, we propose a combined analysis that looks at MSE and TED together. In this case, an MSE value that will be considered a solution's success must be chosen, to filter out expressions that present small errors despite presenting no symbolic meaning. This kind of analysis should be adapted to error values appropriate to one's problem. Nonetheless, common ranges usually considered as stopping conditions (such as $10^{-6}$) will represent a great range of scenarios, especially given that every method is subject to the same criteria. Besides, given that the model has reached the expected solution, tuning the parameters for smaller errors is easy enough. Such considerations are not strange to SR analysis, given that some success rate is usually applied to these studies \cite{de2015evaluating}, in which a fixed value for MSE is commonly chosen. Such a process is similar to selecting the number of bins in a histogram, it carries an intrinsic arbitrariness, and different analyses will surface for distinct values, nonetheless, one must be chosen.

\begin{figure}[h!]
	\centering
	\includegraphics[width=\linewidth]{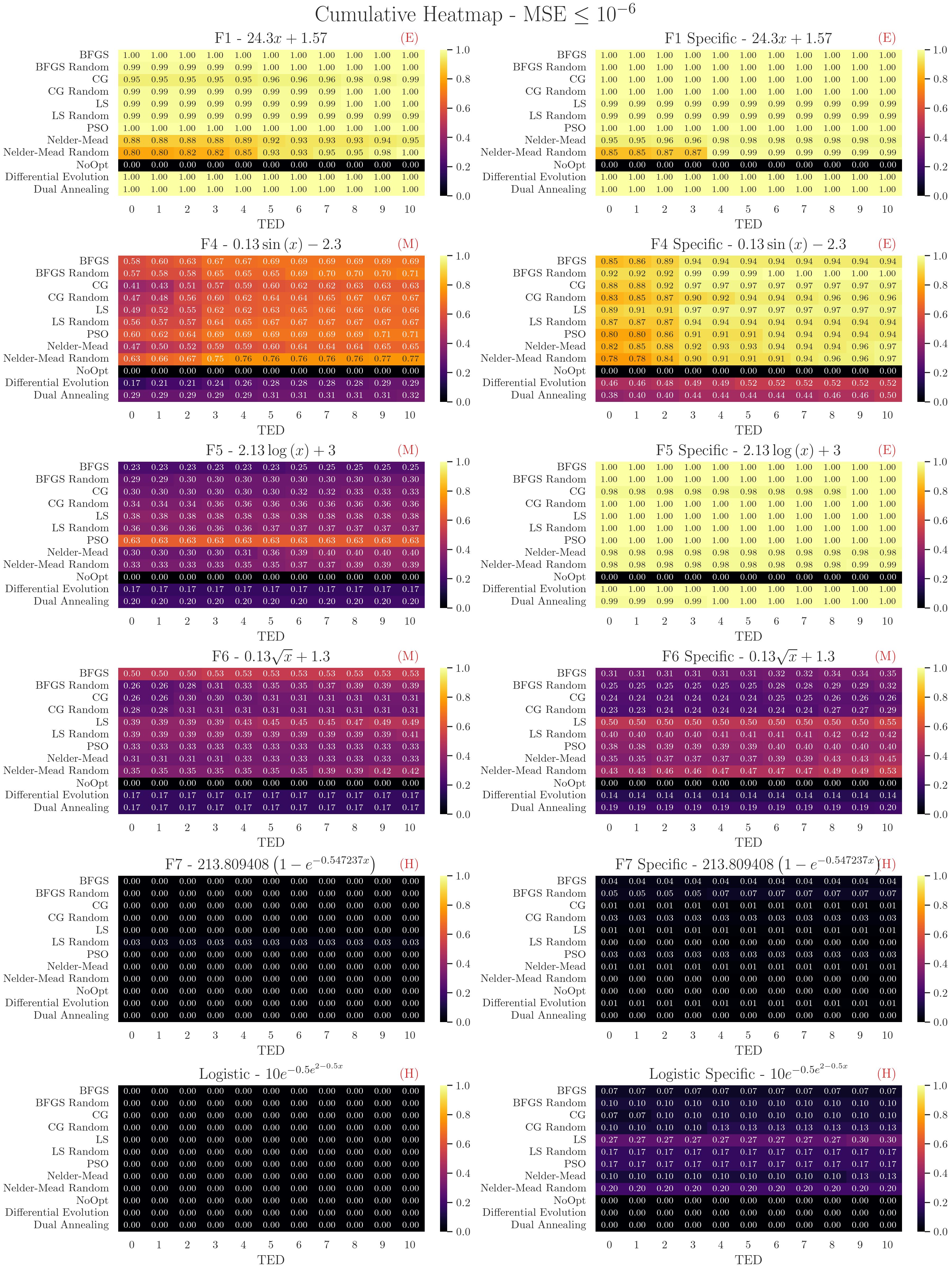}
	\caption{Cumulative heatmap for $MSE \leq 10^{-6}$. Each row represents an optimization method and each step is the number of solutions that have reached the respective value of TED, for the given MSE.}
	\label{fig:cumulative_distribution}
\end{figure}

In \cref{fig:cumulative_distribution}, we present a cumulative distribution plot for a range of TED values, and MSE less than or equal to $10^{-6}$. The horizontal axis represents the value of TED for which the number of points converges. First, it is noticeable that NoOpt did not converge for any threshold in all cases studied. This is the desired behavior, given that this method is incapable of optimizing constants. As discussed previously, it only reaches low values of TED because the given solutions are short. Thus, we were successful in removing this odd behavior, using the analysis in \cref{fig:cumulative_distribution}. Also, for most problems and methods, the distribution is concentrated at $\text{TED} = 0$, which could show interesting behavior concerning the convergence of symbolic regression models.

Secondly, as expected, similar results for the easy cases are observed. Solutions appear to be independent of the constant optimization chosen, given that at least one of them is applied. Following, harder problems also exhibit distributions that appear to be method-independent, where for the majority of cases no algorithm is capable of leading the SR model to the right solution. This behavior seems to indicate that the symbolic regression algorithm utilized is not capable of accurately solving these problems, and any values different than zero are probably caused by chance. The only exception is the \textit{logistic specific} problem, for which case the given information significantly improved performance. \textit{LS}, \textit{LS Random}, \textit{Nelder-Mead Random} and \textit{PSO} achieve some success compared to the other methods, even though it is not as expressive as in different cases. Interestingly such information was not easy to gather from the TED or MSE distributions alone, back in \cref{fig:MSE_boxplot} and \cref{fig:TED_boxplot}.

For the medium cases, it becomes clearer to understand which methods performed better in each problem. For instance in F4, \textit{BFGS}, \textit{PSO}, and especially \textit{Nelder-Mead Random} performed slightly better than other methods. Additionally, in F5, \textit{PSO} has greatly overperformed the alternative algorithms, just like \textit{BFGS} on F6 and \textit{LS} and \textit{Nelder-Mead Random} for F6 Specific. It is curious to notice that even though \textit{BFGS} performed the best in F6, it could not converge for the specific case, that should be easier, given that more information is provided. This could imply that this method is more capable of dealing with problems without previous information.

\begin{figure}[h!]
	\centering
	\includegraphics[width=\linewidth]{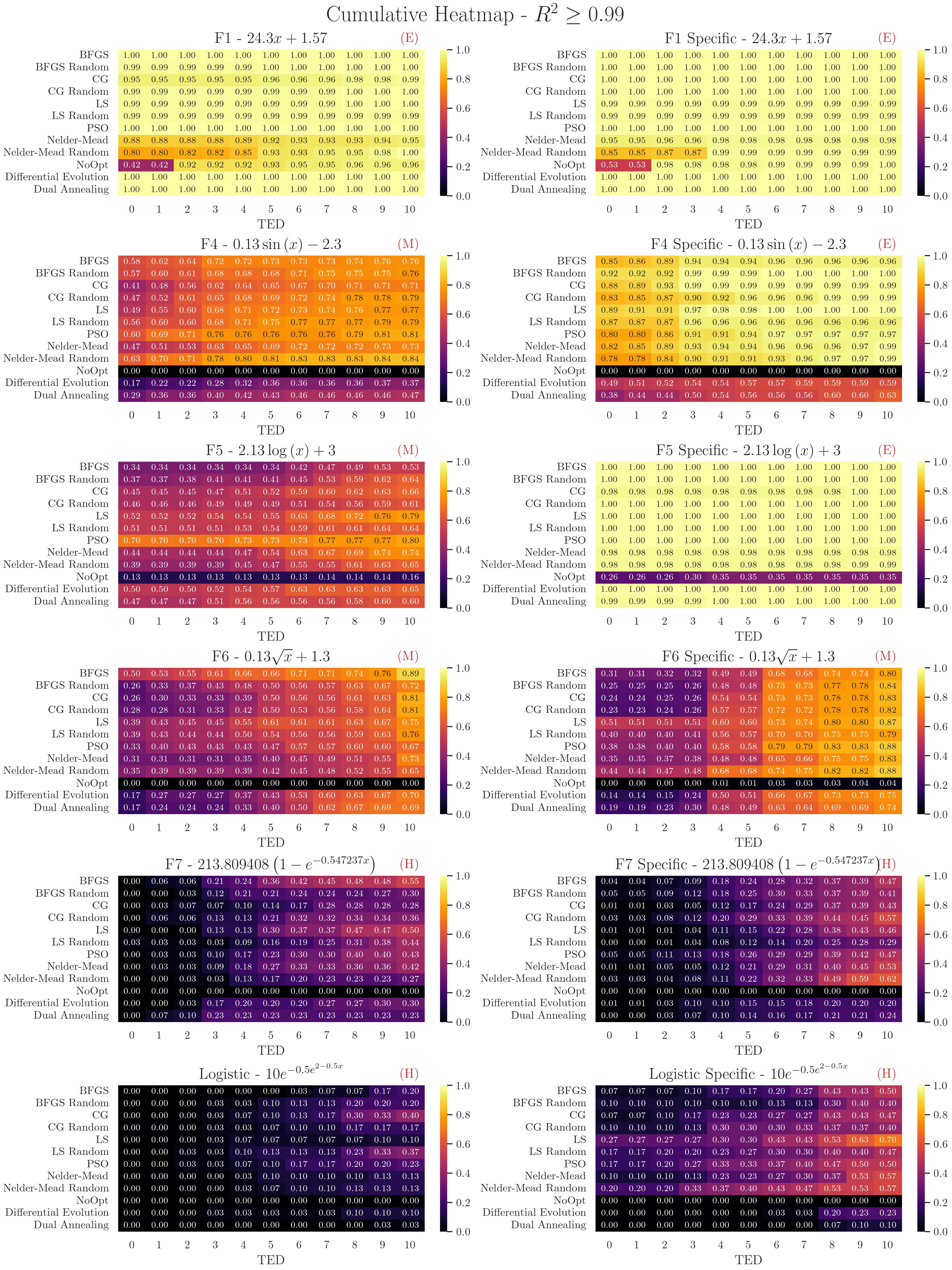}
	\caption{Cumulative heatmap for $R^2 \geq 0.99$. Each row represents an optimization method and each step is the number of solutions that have reached the respective value of TED, for the given $R^2$.}
	\label{fig:cumulative_distribution_R2}
\end{figure}

On the same note, \cref{fig:cumulative_distribution_R2} shows the cumulative heatmap concerning $R^2 > 0.99$. Overall, the same behavior from \cref{fig:cumulative_distribution} is observed for all types of problems. Optimization methods that have outperformed the other ones in the previous case, continue to show better results now. The major difference in the $R^2$ distribution is its concentration. $\text{MSE}$ cumulative plots show results very concentrated in $TED = 0$, and values that don't change much as this threshold increases. On the other hand, $R^2$ heatmap shows significant variation for different $\text{TED}$ values in every problem, except the easy ones. 

We also note that in some cases, such as $F1$ and $F5$, $NoOpt$ also reaches meaningful results, an aspect not observed in \cref{fig:cumulative_distribution}. Furthermore, even though the threshold for $R^2$ is very high, increasing $TED$ meaningfully improves the number of "good" solutions. This is especially noticeable in harder problems, for which the distribution is concentrated around a higher value of TED. Such behaviors imply that most SR solutions can explain data variation, especially in larger expressions. However, it does not indicate how symbolically accurate these expressions are, as represented in figure \cref{fig:size_correlation}. Whereas for $\text{MSE}$, it appears that a high correlation is observed between zero $TED$ and low error. This elucidates an expected behavior, given that only if the model has found the exact expression it is capable of minimizing point-wise comparison.

\bgroup
\newcolumntype{L}{>{\raggedright\arraybackslash}p{1.8cm}}
\newcolumntype{M}{>{\centering\arraybackslash}p{1.5cm}}
\begin{table}[h!]
	\centering
	\begin{tabular}{LrrrMr}
		\toprule
		& MSE & R2 & TED & Training Time (s) & Size \\
		\midrule
		BFGS & 6.43 & 0.91 & 4.59 & 3561.56 & 10.49 \\[2mm]
		BFGS Random & 7.03 & 0.90 & 4.84 & 3656.88 & 10.70 \\[2mm]
		CG & 6.69 & 0.91 & 4.73 & 3657.76 & 10.59 \\[2mm]
		CG Random & 7.21 & 0.91 & 4.60 & 3667.61 & 10.48 \\[2mm]
		LS & 6.47 & 0.91 & 4.43 & 3903.88 & 10.28 \\[2mm]
		LS Random & 6.31 & 0.91 & 4.84 & 3733.46 & 10.64 \\[2mm]
		PSO & 5.14 & 0.93 & 4.23 & 6785.12 & 10.23 \\[2mm]
		Nelder-Mead & 6.90 & 0.91 & 4.79 & 3814.89 & 10.47 \\[2mm]
		Nelder-Mead Random & 6.78 & 0.91 & 4.13 & 3656.49 & 9.82 \\[6mm]
		NoOpt & 117.01 & 0.69 & 4.88 & 4076.70 & 7.41 \\[2mm]
		Differential Evolution & 10.15 & 0.86 & 4.37 & 4123.04 & 9.97 \\[6mm]
		Dual Annealing & 9.97 & 0.87 & 4.34 & 4706.08 & 9.92 \\
		\bottomrule
	\end{tabular}
	\caption{Average values over every problem for of each metric studied and every optimization algorithm.}
	\label{tab:average_table}
\end{table}
\egroup

In \cref{tab:average_table} we see the average value for each metric studied and every optimization algorithm. It should be noted that, even though this table summarizes the performance of each method, it does not contain every behavior discussed so far. This aspect reveals the importance of looking closer, at every test problem, to understand not-so-obvious behaviors. Despite that, the summarized results do show some important aspects we have seen so far.

First, there's no clear winner between random and standard versions. It appears that for which side this relationship tends to depends on the method chosen, and in different intensities. Second, very similar training times are observed among every choice. As exceptions, we must highlight PSO and Dual Annealing. The first one is probably explained by the fact that PSO is the only algorithm not implemented by Python's \textit{scipy} library (that has Fortran background). Instead, we utilize \textit{pyswarms}, which is entirely written in Python. Dual Annealing on the other hand is already known as a slower method, thus this outcome is not surprising.

Concerning expression size, every optimization algorithm reaches very similar averages. The only exception is \textit{NoOpt}, which presents an average size significantly smaller. This outcome agrees with the conclusion we have reached before. It also explains why this method presents average values for TED, even though it is not capable of actually optimizing the solution. Finally, even though only by a small margin, PSO and Nelder-Mead Random presented themselves as the best options. They appear to be suited as default optimization algorithms in symbolic regression models.

\section{Conclusion}
\label{sec:conclusions}
In this work, we have studied the impact different constant optimization methods have on Symbolic Regression predictions when applied during the evolutionary search. Nine test problems were studied in two different scenarios, with varying amounts of information. To evaluate performance, we utilized $\text{MSE}$ and $R^2$ in conjunction with a new metric called $\text{TED}$ to estimate the symbolic accuracy of found expressions. A preprocessing simplification procedure was also implemented, to achieve more accurate results.

Investigation of $\text{MSE}$ and $\text{TED}$ distributions separately has led us to identify three major groups of problems. For easy ones, every approach converges, given that a constant optimization method is chosen. In hard problems, no alternative converges, indicating that the issue is not related to parameter optimization. At last, medium problems exhibit behavior that is method-dependent, meaning that for each particular case, different methods perform best. Given that this classification depends on how powerful the SR model is, it means that for most problems we should expect that different constant optimization techniques will produce distinct results. 

Following, we detect a very high correlation between expression size (complexity) and $\text{TED}$. Such behavior implies that smaller solutions produce lower values of $\text{TED}$. We have shown that this behavior relates to inaccurate evaluations and that it is necessary to consider expression size when analyzing $\text{TED}$.

At last, we have proposed a combined analysis of numeric and symbolic errors. Such an approach allows us to understand better the actual percentage of useful/meaningful solutions, that accurately describe a given problem. The analysis revealed that methods like \textit{PSO} and \textit{BFGS} should be set as default in SR models, once they perform better over a wider range of problems. In one scenario, \textit{BFGS} has achieved greater performance with less information. In contrast, \textit{Levenbergâ€“Marquardt} and \textit{Nelder-Mead} have shown to be more powerful alternatives to harder problems. Although, the latter presents a curious behavior, in which it performs poorly in easy to medium problems, and greatly improves performance for harder ones. 

Additionally, studying $R^2$ distributions, we notice that this metric is way more ``forgiving" compared to $\text{MSE}$ and could give the impression of a better method. Many more solutions can describe data variation, but only those that achieve low error seem to produce accurate symbolic expressions. Also note that, contrary to what is standard in the area, a table of average values, over a wide range of problems and iterations, is not capable of accurately describing every behavior observed, and a more thorough analysis is required.

\printbibliography

\end{document}